\documentclass{article}
\usepackage{spconf,amsmath,graphicx}
\usepackage{amsmath}
\usepackage{graphicx}
\usepackage{float}

\usepackage{graphicx}
\usepackage{epsfig}
\usepackage{amsmath}
\usepackage{amssymb,algorithmic}
\usepackage{subfigure}
\usepackage{wrapfig}
\usepackage{times,color}
\usepackage{cite,footnote,xspace,syntonly,theorem,algorithm,algorithmic}
\input{mysymbol.sty}

\title{Graph Filtering For Data Reduction and Reconstruction}
\name{Ioannis D. Schizas}
\address{Department of Electrical Engineering\\
University of Texas at Arlington}
\begin{document}
%\ninept
%
\maketitle
\begin{abstract}
A novel approach is put forth that utilizes data similarity,  quantified on a graph, to improve upon the reconstruction performance of principal component analysis. The tasks of data dimensionality reduction and reconstruction are formulated as graph filtering operations, that enable the exploitation of data node connectivity in a graph via the adjacency matrix. The unknown reducing and reconstruction filters are determined by optimizing a mean-square error cost that entails the data, as well as their graph adjacency matrix. Working in the graph spectral domain enables the derivation of simple gradient descent recursions used to update the matrix filter taps. Numerical tests in real image datasets demonstrate the better reconstruction performance of the novel method over standard principal component analysis.
\end{abstract}
\section{Introduction}
\label{sec:intro}
Data dimensionality reduction and reconstruction has been extensively studied, with the workhorse approach being the principal component analysis (PCA) framework which determines proper compression and reconstruction matrices that minimize the mean-square error (MSE),  see e.g., \cite{Brillinger}. Standard PCA relies on data correlations within each data vector to find a MSE-optimal data representation in a reduced dimensional space. Our goal here, is to exploit similarity among different data vectors when performing dimensionality reduction, manifested as edge weights on a graph, to improve the data reconstruction performance.

Graph signal processing is an emerging field where similarity among the available data is exploited, via the utility of shift operators, to improve the performance in a variety of tasks including sampling, filtering, clustering and sampling/reconstruction \cite{Sampling_1,Tutorial_FFT_Filtering,Graph_Tutorial}. The concept of sampling a graph signal in a subset of nodes and reconstructing it wherever is not available has been extensively explored \cite{Sampling_1,Sampling_2,Time_Varying_Graph_Signal_Rec, Graph_Signal_Rec_Percolation,Wang_Sampling}. In these works, the idea of bandlimited signals is extended in the graph spectral domain, and techniques exploiting the Laplacian eigenspace are devised to reconstruct the signal values in every node of the graph from a subset of nodes.

Dimensionality reduction in graphs has been proposed by expanding the PCA or nonnegative matrix factorization formulations with a Laplacian regularization term that takes into account similarity among single-hop neighboring data entities in a graph  \cite{Multiple_Graph_Coding,Graph_NMF,Graph_Lapl_PCA,Shen_GG,Robust_PCA_Graphs}. In the aforementioned line of work dimensionality reduction is performed to improve data clustering performance. Differently, our goal here is data dimensionality reduction and reconstruction  by exploiting data similarity quantified here by the graph adjacency matrix. 

The tasks of data dimensionality reduction and reconstruction are carried out via graph filtering, while the order of the matrix filters will determine the neighborhood size that will be utilized in determining the compressed and reconstructed data. The novel formulation is seeking MSE-optimal filter matrices that minimize the reconstruction MSE in the graph. A computationally effective gradient descent approach is proposed to recursively determine the filters. For zero-order filters the novel framework boils down to standard PCA. Numerical tests using real image datasets demonstrate the superiority of the novel graph-based dimensionality reduction and reconstruction framework over standard PCA. 

\section{Problem Setting and Preliminaries}
\label{sec:problem_statement}
Consider a collection of data $\bbX:=[\bbx_1\ldots\bbx_n]$, where  each data vector $\bbx_i$ has  $D$ scalar entries.
Columns in $\bbX$ could correspond to a collection of images, sensor measurements and so on \cite{EYB, Graph_NMF}. In many practical applications the data vectors lie on a low dimensional vector space $\mathbb{R}^{d\times 1}$, where $d<<D$.

One of the most effective ways to apply dimensionality reduction to the data is to employ principal component analysis (PCA), see e.g., \cite{Brillinger}. PCA, being the dimensionality reduction workhorse, extracts the principal components by projecting the data onto a low dimensional vector subspace in which the data demonstrate the largest variability.
PCA is determining a dimensionality reducing matrix $\hat{\bbC}$ of size $k\times D$, with $k\leq D$ and a reconstruction matrix $\hat{\bbB}\in\mathbb{R}^{D\times k}$, which are found by minimizing the reconstruction MSE 
\begin{equation}\label{Eq:PCA}
\{\hat{\bbB},\hat{\bbC}\}=\arg\min_{\bbB,\bbC}n^{-1}\|\bar{\bbX}-\bbB\bbC\bar{\bbX}\|_F^2,
\end{equation}
where $\bar{\bbX}:=[\bar{\bbx}_1\ldots\bar{\bbx}_n]$ corresponds to a centered version of the data, with  $\bar{\bbx}_i:=\bbx_i-n^{-1}\sum_{j=1}^{n}\bbx_j$ for $j=1,\ldots,n$, and $\|\cdot\|_F$ denotes the Frobenius norm. It turns out that $\hat{\bbC}^T=\hat{\bbB}=\bar{\bbU}_{x,k}$, where $\bar{\bbU}_{x,k}$
 contains in its columns the $k$ principal eigenvectors of sample-average covariance matrix $\bar{\bbSigma}_x:=n^{-1}\bar{\bbX}\bar{\bbX}^T$.

PCA is designed to estimate the low dimensional subspace $\bar{\bbU}_{x,k}$ using $\bar{\bbSigma}_x$, without taking into account similarity among different data vectors. However, the dataset $\bbX$ may contain groups of data vectors that exhibit similarity in some sense, e.g., images depicting a similar object or having similar texture. Standard PCA does not take into account data similarity information that can potentially identify structurally similar data and lead to better reconstruction.
 
Data similarity measures if available can be utilized in a graph. Specifically, let scalar $s_{ij}$ quantify the similarity between data vectors $\bbx_i$ and $\bbx_j$ for $i,j=1,\ldots,n$. Then, an undirected graph ${\cal G}$ with $n$ nodes within set ${\cal V}:=\{1,\ldots,n\}$ and edges in ${\cal E}:=\{(i,j):\;s_{ij}\neq 0 \textrm{ for  } i,j=1,\ldots,n\}$ can summarize the similarity among the different data in $\bbX$. Note that since the graph is undirected then $s_{ij}=s_{ji}$. The similarity quantities can be summarized in the so called adjacency matrix $\bbS:=[s_{ij}]$ which is an $n\times n$ symmetric matrix whose eigenvalue decomposition can be written as $\bbS=\bbU\bbLambda\bbU^T$, where $\bbLambda:=\textrm{diag}(\lambda_1,\ldots,\lambda_n)$ is a $n\times n$ diagonal matrix that contains the eigenvalues, while $\bbU:=[\bbu_{1}\ldots\bbu_{n}]$ is a unitary matrix containing the eigenvectors of $\bbS$.

PCA is redesigned in this work to exploit data similarities summarized in the adjacency matrix $\bbS$, via graph filtering, and improve reconstruction performance.

\section{Data Reduction and Reconstruction via Graph Filtering}

To exploit the similarity weights on the graph edges we utilize graph filtering (GF) \cite{Tutorial_FFT_Filtering,Graph_Signal_Rec_Percolation,Graph_Tutorial}. A scalar linear shift-invariant graph filter of order $L$ is given as, see e.g., \cite{Tutorial_FFT_Filtering,Graph_Signal_Rec_Percolation,Graph_Tutorial} 
\begin{equation}\label{Eq:GFS}
\bbK:=\textstyle\sum_{\ell=0}^{L}c_{\ell}\bbS^{\ell},
\end{equation}
%%
%%
%while for input $\bbX$ with $D=1$ (scalar signal per graph node) the output is $\bbY=\sum_{\ell=0}^{L}c_{\ell}\bbS^{\ell}$, 
where $\bbS$ denotes a graph shift operator that in this paper will be the adjacency matrix $\bbS$. Building upon \eqref{Eq:GFS}
we define the following data reducing graph matrix filtering operation
\begin{equation}\label{Eq:GFS_RD}
\bby:=\textstyle\sum_{\ell=0}^{L}(\bbS^{\ell}\otimes \bbI_{D})\cdot (\bbI_{n}\otimes \bbC_{\ell})\breve{\bbx},
\end{equation}
where $\breve{\bbx}:=\textrm{vec}(\bar{\bbX})\in\mathbb{R}^{nD\times 1}$ is obtained after stacking the columns in $\bar{\bbX}$ on top of each other, while $\bbI_{n}$ refers to an identify matrix of size $n\times n$ and $\otimes $ is the Kronecker product. 

Vector $\bby:=[\bby_1^T\ldots\bby_n^T]\in\mathbb{R}^{nk\times 1}$ contains the reduced dimensionality data vectors $\bby_i$ with $k$
entries for each node $i=1,\ldots,n$, while each $\bby_i$  is produced by compressing and linearly combining data vectors from neighboring nodes (up to $L$ hops away from node $i$) using the dimensionality reducing matrices $\bbC_{\ell}\in\mathbb{R}^{k\times D}$ for $\ell=0,\ldots,L$. The motivation behind this reducing filtering step is that data vectors within a neighborhood of few hops will exhibit large similarity, and these data can be used jointly to better reduce to $\bby_i$ the contents of  $\bar{\bbx}_i$. Note that for $L=0$, \eqref{Eq:GFS_RD} boils down to $\bby_i=\bbC_0\cdot\bar{\bbx}_i$ which pertains to standard PCA.

Similarly, graph filtering can be utilized as in \eqref{Eq:GFS_RD} to reconstruct the data vectors using the reduced vectors $\bby_i$, the adjacency matrix $\bbS$ and reconstruction matrices $\bbB_{\ell}\in\mathbb{R}^{D\times k}$ in the following way
\begin{equation}\label{Eq:GFS_Rec}
\hat{\bbx}:=\textstyle\sum_{m=0}^{L}(\bbS^{m}\otimes \bbI_{D})\cdot (\bbI_{n}\otimes \bbB_{m}){\bby}.
\end{equation}

The dimensionality reducing and reconstruction matrices $\{\bbB_{\ell},\bbC_{\ell}\}_{\ell=0}^{L}$ will be determined such that the reconstruction MSE resulting after applying \eqref{Eq:GFS_RD} and \eqref{Eq:GFS_Rec} is minimized, i.e.,
\begin{align}\label{Eq:GPCA}
\{\hat{\bbB}_{\ell},\hat{\bbC}_{\ell}\}_{\ell=0}^{L}:=\arg\min_{\bbB_{\ell},\bbC_{\ell}}&n^{-1}\|\breve{\bbx}-
\textstyle\sum_{m=0}^{L}(\bbS^{m}\otimes \bbI_{D})\nonumber\\
&\hspace{-3cm}\cdot (\bbI_{n}\otimes \bbB_{m})\cdot\textstyle\sum_{\ell=0}^{L}(\bbS^{\ell}\otimes \bbI_{D})\cdot (\bbI_{n}\otimes \bbC_{\ell})\breve{\bbx}\|_{2}^2.
\end{align}
For simplicity it has been assumed that the order of the reducing and reconstruction filters is $L$, nonetheless the proposed framework allows for different orders. Note that for $L=0$ the cost function in \eqref{Eq:GPCA} boils down to $n^{-1}\sum_{i=1}^n\|\bar{\bbx}_i-\bbB_0\bbC_0\bar{\bbx}_i\|_2^2$ which corresponds to the standard PCA formulation which does not take into account data similarity information.

\subsection{Graph Spectrum MSE Reformulation }
The cost function in \eqref{Eq:GPCA} is reformulated next to facilitate the determination of the matrix filter taps
$\{\hat{\bbB}_{\ell},\hat{\bbC}_{\ell}\}_{\ell=0}^{L}$. %Note that the number of scalar optimization variables in \eqref{Eq:GPCA} is proportional to the dimensionality of the data $D$.
Multiplication of $\breve{\bbx}$ and $\hat{\bbx}$ in \eqref{Eq:GPCA} with the unitary matrix $\bbU_{\alpha}^T:=\bbU^T\otimes \bbI_{D}$ has no effect in the cost, i.e., $\|\breve{\bbx}-\hat{\bbx}\|_2^2=\|\bbU_{\alpha}^T(\breve{\bbx}-\hat{\bbx})\|_2^2$. Let $\tilde{\bbx}:=[\tilde{\bbx}_1^T\ldots\tilde{\bbx}_n^T]^T=\bbU_{\alpha}^T\breve{\bbx}$ denote the graph Fourier transform (GFT) of the data $\breve{\bbx}$ with respect to the adjacency matrix $\bbS$. In detail, the GFT at the ith frequency (ith eigenvalue of $\bbS$) is given as
$\tilde{\bbx}_i=\sum_{j=1}^{n}\bbU(i,j)\bar{\bbx}_i$, where $\bbU(i,j)$ corresponds to the $(i,j)$th entry of $\bbU$. After the unitary transformation of the reconstruction MSE and using the property that $\bbS^{\ell}=\bbU\bbLambda^{\ell}\bbU^T,$  for $\ell=0,\ldots,L$, the minimization problem in \eqref{Eq:GPCA} can be rewritten as
\begin{equation}\label{Eq:GPCA_GFT}
\{\hat{\bbB}_{\ell},\hat{\bbC}_{\ell}\}_{\ell=0}^{L}:=\arg\min_{\bbB_{\ell},\bbC_{\ell}}n^{-1}\sum_{i=1}^{n}\|\tilde{\bbx}_i-\bbW_i\bbB\bbC\bbW_i^T\tilde{\bbx}_i\|_2^2,
\end{equation}
where $\bbW_i:=[\bbI_D\;\lambda_i\bbI_{D}\;\ldots\lambda_i^L\bbI_{D}]$, while $\bbB:=[\bbB_0^T\ldots\bbB_{L}^T]^T$ and 
$\bbC:=[\bbC_0\ldots\bbC_{L}]$.  Thus, \eqref{Eq:GPCA_GFT} can be viewed as a spectral version of \eqref{Eq:GPCA} and convolution has been transformed into a multiplication between the filters' spectral response and the GFT of the data vectors. 
Note that $\tilde{\bbB}_{i}:=\bbW_i\bbB=\sum_{\ell=0}^{L}\lambda_{i}^{\ell}\bbB_{\ell}$ can be viewed as the spectral response of the reconstruction matrix filter $\bbB$ at eigenvalue $\lambda_i$, similarly $\bbC\bbW_i^T$ corresponds to the spectral response of the reducing matrix filter at $\lambda_i$.

The cost function in \eqref{Eq:GPCA_GFT} can be rewritten as follows
\begin{align}\label{Eq:GPCA_GFT_v2}
&\textrm{tr}(\bbSigma_x)-2\textrm{tr}(\bbB\bbC\bbSigma_{\tilde{z}})\\
&+n^{-1}\textstyle\sum_{i=1}^{n}\textrm{tr}(\bbW_i\bbB\bbC\tilde{\bbz}_i\tilde{\bbz}_i^T\bbC^T\bbB^T\bbW_i^T)\nonumber
\end{align}
where $\tilde{\bbz}_i:=\bbW_i^T\tilde{\bbx}_i$ and $\bbSigma_{\tilde{z}}:=n^{-1}\sum_{i=1}^{n}\tilde{\bbz}_i\tilde{\bbz}_i^T$.

Taking first-order derivatives of \eqref{Eq:GPCA_GFT_v2} with respect to (wrt) $\bbC$ and $\bbB$ and setting them equal to zero, we obtain the following first-order optimality conditions \cite{Bertsekas_Nonlinear_Book}
\begin{align}\label{Eq:Diff}
\bbC\bbSigma_{\tilde{z}}&=\bbC \left[n^{-1}\textstyle\sum_{i=1}^n\tilde{\bbz}_i\tilde{\bbz}_i^T\bbC^T\bbB^T\bbW_i^T\bbW_i\right],\\
\bbSigma_{\tilde{z}}\bbB&=\left[n^{-1}\textstyle\sum_{i=1}^n\tilde{\bbz}_i\tilde{\bbz}_i^T\bbC^T\bbB^T\bbW_i^T\bbW_i\right]\bbB.
\nonumber
\end{align}
The equalities in \eqref{Eq:Diff} can be utilized to show the following result (the proof has been omitted due to space considerations).
\corollary The reducing matrix filter taps in $\bbC$ can be written as a linear combination of the transformed  data vectors $\tilde{\bbz}_i=\bbW_i^T\tilde{\bbx}_i$, i.e., 
\begin{equation}\label{Eq:BCY}
\bbC=\bbG\cdot\tilde{\bbZ}^T,
%\bbB=\tilde{\bbZ}\cdot\bbF,\;\;
\end{equation}
where $\tilde{\bbZ}:=[\tilde{\bbz}_1\ldots\tilde{\bbz}_n]$, while  $\bbG\in\mathbb{R}^{k\times n}$.\normalfont\newline

The result of Corollary 1 can be utilized to replace $\bbC$ with $\bbG$ in \eqref{Eq:GPCA_GFT} reducing in that way the number of primary optimization variables. Note that $\bbC$ contains $k(L+1)D$ entries that need to be found, whereas $\bbG$ has $kn$ entries that need to be determined.
For applications where $D>>n$, Cor. 1 can be used to introduce computational savings when solving \eqref{Eq:GPCA_GFT}. 

\subsection{Gradient Descent Based Algorithm}

We resort to a gradient descent approach to devise a computationally simpler method to minimize the cost in \eqref{Eq:GPCA_GFT_v2}. Specifically, during iteration $\kappa+1$ the gradient descent updates \cite{Bertsekas_Nonlinear_Book} for $\bbB$ and $\bbG$ are given as 
\begin{equation}\label{Eq:Grd_Desc}
\bbB^{\kappa+1}=\bbB^{\kappa}-c_B^{\kappa+1}\nabla\bbB^{\kappa},\;\;
\bbG^{\kappa+1}=\bbG^{\kappa}-c_G^{\kappa+1}\nabla\bbG^{\kappa},
\end{equation}
where $c_B^{\kappa+1},c_G^{\kappa+1}$ are nonnegative step-sizes to be determined by line-search later on, and $\nabla\bbB^{\kappa}$, $\nabla\bbG^{\kappa}$ are the gradients of the cost function in  \eqref{Eq:GPCA_GFT_v2}  evaluated wrt  $\bbB$ and $\bbG$, respectively. Differentiation of \eqref{Eq:GPCA_GFT_v2}  wrt $\bbB$ gives
\begin{eqnarray}\label{Eq:Nabla_B}
\nabla\bbB^{\kappa}&=&n^{-1}\textstyle\sum_{i=1}^n\bbW_i^T\tilde{\bbB}_i^{\kappa}\bbG^{\kappa}\bbK_{z,:i}\bbK_{z,:i}^T(\bbG^{\kappa})^T-\\
&&\textstyle n^{-1}\sum_{i=1}^n\bbW_i^T\tilde{\bbx}_i\bbK_{z,:i}^T(\bbG^{\kappa})^T\nonumber\\
&=&n^{-1}\textstyle\sum_{i=1}^{n}\bbW_i^T\left(\tilde{\bbx}_i-\tilde{\bbB}_i^{\kappa}\bbG^{\kappa}\bbK_{z,:i}\right)\bbK_{z,:i}^T(\bbG^{\kappa})^T\nonumber
\end{eqnarray}
where $\bbK_{z}:=\tilde{\bbZ}^T\tilde{\bbZ}$ and $\bbK_{z,:i}$ denotes the $i$th column of $\bbK_z$, i.e., $\bbK_{z,:i}=\tilde{\bbZ}^T\tilde{\bbz}_{i}$ and $\tilde{\bbB}_i^{\kappa}=\bbW_i\bbB^{\kappa}$.

Similarly, the gradient $\nabla\bbG^{\kappa}$ can be calculated as
\begin{equation}\label{Eq:Nabla_G}
\nabla\bbG^{\kappa}=n^{-1}\textstyle\sum_{i=1}^{n}(\tilde{\bbB}_i^{\kappa+1})^T(\tilde{\bbx}_i-\tilde{\bbB}_i^{\kappa+1}\bbG^{\kappa}\bbK_{z,:i})\bbK_{z,:i}^T.\nonumber
\end{equation}

From \eqref{Eq:Grd_Desc} and \eqref{Eq:Nabla_B} each $D\times k$ submatrix $\bbB_{\ell}$ in $\bbB$ for $\ell=0,\ldots,L$ can be updated as
\begin{equation}\label{Eq:Bell_update}
\bbB_{\ell}^{\kappa+1}=\bbB_{\ell}^{\kappa}-c_{B}^{\kappa+1}n^{-1}\textstyle\sum_{i=1}^{n}\lambda_{i}^{\ell}(\tilde{\bbx}_i-\tilde{\bbB}_i\bbG^{\kappa}\bbK_{z,:i})\bbK_{z,:i}^T({\bbG^{\kappa}})^T,
\end{equation}
whereas $\bbG^{\kappa+1}$ is found as 
\begin{equation}\label{Eq:G_Update}
\bbG^{\kappa+1}=\bbG^{\kappa}-c_G^{\kappa+1}n^{-1}\textstyle\sum_{i=1}^{n}\tilde{\bbB}_i^T(\tilde{\bbx}_i-\tilde{\bbB}_i\bbG^{\kappa}\bbK_{z,:i})\bbK_{z,:i}^T.
\end{equation}
%%

 %%Talk about complexity
The computational complexity (number of additions and multiplications) for carrying out the the gradient descent recursions in \eqref{Eq:Bell_update}  is of the order of ${\cal O}(k(L+1)(Dn+n^2))$, while for \eqref{Eq:G_Update} complexity is of the order of 
${\cal O}(k(Dn+n^2))$. Complexity is proportional to the dimensionality of the data vectors $D$, the order of the filters $L$ and quadratic in $n$. 

\noindent \textbf{Optimal step-size selection:} We resort to line search, see e.g.,  \cite{Bertsekas_Nonlinear_Book}, where the step-sizes in $c_B^{\kappa}$ and $c_G^{\kappa}$ are set such that they minimize the cost function in \eqref{Eq:GPCA_GFT} after substituting $\bbB$ and $\bbG$ with the updating recursions in \eqref{Eq:Bell_update} and  \eqref{Eq:G_Update} and minimizing wrt to the $c_B$ or $c_G$ parameters. We demonstrate the process for $c_B$. After substituting $\bbB$  in \eqref{Eq:GPCA_GFT_v2} with the right hand side in \eqref{Eq:Bell_update}, and $\bbC=\bbG\tilde{\bbZ}^T$ it turns out that the optimal choice for $c_{B}$ during iteration $\kappa+1$ can be obtained as
\begin{align}\label{Eq:Find_CB}
\hspace{-0.15cm}&c_{B}^{\kappa+1}=\arg\min_{c}-2\cdot\textrm{tr}[(\bbB^{\kappa}-c\nabla\bbB^{\kappa})\bbG^{\kappa}\tilde{\bbZ}^T\bbSigma_{\tilde{z}}]+n^{-1}\textstyle\sum_{i=1}^{n}\nonumber\\
\hspace{-0.15cm}&\textrm{tr}[\bbW_i(\bbB^{\kappa}-c\nabla\bbB^{\kappa})\bbG^{\kappa}\bbK_{z,:i}\bbK_{z,:i}^T(\bbG^{\kappa})^T(\bbB^{\kappa}-c\nabla\bbB^{\kappa})^T\bbW_i^T]\nonumber\\
\hspace{-0.15cm}&=\arg\min_{c}2c\gamma_{1}^{\kappa}-2c\gamma_{2}^{\kappa}+c^2\gamma_3^{\kappa},
\end{align}
where $\gamma_1^{\kappa}:=n^{-1}\sum_{i=1}^n\tilde{\bbx}_i^T\nabla\tilde{\bbB}_{i}^{\kappa}\bbG^{\kappa}\bbK_{z,:i}$ with $\nabla\tilde{\bbB}_{i}:=\sum_{\ell=0}^{L}\lambda_i^{\ell}\nabla\bbB_{\ell}^{\kappa}$. Further,
the quantities $\gamma_2^{\kappa}$ and $\gamma_3^{\kappa}$ are
\begin{align}\label{Eq:Gamma23}
\hspace{-0.15cm}&\gamma_2^{\kappa}:=n^{-1}\textstyle\sum_{i=1}^{n}\bbK_{z,:i}^T
(\bbG^{\kappa})^T(\tilde{\bbB}_i^{\kappa})^T\nabla\tilde{\bbB}_i^{\kappa}\bbG^{\kappa}\bbK_{z,:i},\\
\hspace{-0.15cm}&\gamma_3^{\kappa}:=n^{-1}\textstyle\sum_{i=1}^{n}\bbK_{z,:i}^T(\bbG^{\kappa})^T(\nabla\tilde{\bbB}_i^{\kappa})^T\nabla\tilde{\bbB}_i^{\kappa}\bbG^{\kappa}\bbK_{z,:i}.
\end{align}
Then, it follows readily that the optimal step-size in \eqref{Eq:Find_CB} is equal to 
$c_{B}^{\kappa+1}=(\gamma_3^{\kappa})^{-1}\cdot(\gamma_2^{\kappa}-\gamma_1^{\kappa})$.

Using a similar approach where we substitute $\bbC$ with $\bbG\tilde{\bbZ}^T$, and then replace $\bbG$ with the right hand side of \eqref{Eq:G_Update} in \eqref{Eq:GPCA_GFT_v2} we can find the optimal selection for step-size $c_G$ as
%%
%\begin{equation}\label{Eq:Opt_step}
$c_G^{\kappa+1}=({\delta_3^{\kappa}})^{-1}({\delta_{2}^{\kappa}-\delta_{1}^{\kappa}}),$
%\end{equation}
%%
where 
\begin{align}\label{Eq:}
&\hspace{-0.15cm}\delta_1^{\kappa}:=n^{-1}\textstyle\sum_{i=1}^{n}\tilde{\bbx}_i^T\tilde{\bbB}_{i}^{\kappa}\nabla\bbG^{\kappa}\bbK_{z,:i},\\
&\hspace{-0.15cm}\delta_2^{\kappa}:=n^{-1}\textstyle\sum_{i=1}^{n}\bbK_{z,:i}^T
(\nabla\bbG^{\kappa})^T(\tilde{\bbB}_i^{\kappa})^T\tilde{\bbB}_i^{\kappa}\bbG^{\kappa}\bbK_{z,:i},\nonumber\\
&\hspace{-0.15cm}\delta_3^{\kappa}:=n^{-1}\textstyle\sum_{i=1}^{n}\bbK_{z,:i}^T(\nabla\bbG^{\kappa})^T(\tilde{\bbB}_i^{\kappa})^T\nabla\tilde{\bbB}_i^{\kappa}\nabla\bbG^{\kappa}\bbK_{z,:i}.\nonumber
\end{align}

\noindent \textbf{Initialization:} $\bbB$ and $\bbG$ can be initialized using the solution of standard PCA to which our framework boils down to when $L=0$. Let the standard PCA compression and reconstruction matrices be denoted as $\hat{\bbB}_0=\hat{\bbC}_0^T=\bar{\bbU}_{x,k}$ Then, 
we can initialize $\bbB$ as $\bbB^0=[\hat{\bbB}_0^T \;\mathbf{0}_{k\times D}\ldots \mathbf{0}_{k\times D}]^T$.  From Corollary 1 it holds that $\bbC_0=\bar{\bbU}_{x,k}^T=\bbG_0\cdot [\tilde{\bbx}_1\ldots\tilde{\bbx}_n]^T=\bbG_0\tilde{\bbX}^T$ (when $L=0$) from which we can obtain  $\bbG^0=\bbC_0\tilde{\bbX}(\tilde{\bbX}^T\tilde{\bbX})^{-1}$. The gradient descent based approach is tabulated as Alg. 1. $\bbG^{\kappa}$ and $\bbB^{\kappa}$ are updated until the norm of the difference between successive iterates drops below a desired threshold $\epsilon$.

\noindent \textbf{Remark:} Note that the original data consist of $n\cdot D$ scalars, which can be prohibitively large. When, applying the dimensionality reduction matrix filter $\bbC$ each data vector is described by $k$ scalars corresponding to the entries of $\{\bby_j\}_{j=1}^{n}$. Thus, a total of $n\cdot k\leq n \cdot D$ scalars are utilized to characterize the dimensionality reduced data. Notice that to form the reconstructed data $\hat{\bbx}$  in (4), the  $k(L+1)D+kn$ entries of $\bbB$ and $\bbG$, as well as the $n(n+1)/2$ different entries of the symmetric adjacency matrix $\bbS$ and the $nk$ scalars in $\bby$ are needed.  The cost of storing  the $k(L+1)D+2kn+n(n+1)/2$ entries of $\bbB,\bbG$, $\bbS$ and $\{\bby_j\}_{j=1}^{n}$ for the graph-based data reduction scheme, is higher than storing $kD+kn$ scalars required in standard PCA for $\hat{\bbB}_0=\hat{\bbC}_0^T$ and the $\{\bby_j\}_{j=1}^{n}$.  Nonetheless, the graph-based approach achieves better reconstruction accuracy as detailed next. Here compression occurs as long as 
\begin{align}\label{Eq:Bound}
&2kn+k(L+1)D+\frac{n(n+1)}{2}\leq nD \\
&\Leftrightarrow k\leq \frac{n[D-0.5(n+1)]}{2n+(L+1)D}.
\end{align}
Thus, for high-dimensional data $D>>n$ (such as images) and a limited amount of data vectors, the right hand side in \eqref{Eq:Bound} can be approximated as $\frac{n}{L+1}$. Thus, as long as $k\leq \frac{n}{L+1}$ there is meaningful data reduction.

%%%%%%Algorithm 1: Kernel Sparse Matrix Decomposition
\begin{algorithm}
	\caption{Gradient Based Matrix Filter Determination} \small
	\begin{algorithmic}[1]
		\itemsep 0.05cm
		\STATE Initiliaze $\bbB^0$ and $\bbG^0$ using standard PCA.
		\FOR {$\kappa=0,1,\ldots$}
		\STATE Determine optimal step-sizes $c_G^{\kappa+1}$ and $c_B^{\kappa+1}$.
		
		\STATE Update $\bbB^{\kappa+1}$ and $\bbG^{\kappa+1}$ via \eqref{Eq:Bell_update} and \eqref{Eq:G_Update}, respectively.
		
		\STATE If $\|\bbG^{\kappa+1}-\bbG^{\kappa}\|_F+\|\bbB^{\kappa+1}-\bbB^{\kappa}\|_F<\epsilon$ then stop.
		\ENDFOR
		
	\end{algorithmic}
\end{algorithm}

\section{Numerical Simulations}
We test and compare the performance of the graph-based reduction and reconstruction approach versus standard PCA (where $L=0$) in the MNIST database of handwritten digits, and the Extended Yale-B (EYB) face image dataset \cite{MNIST, EYB}. The MNIST dataset consists of $28\times 28$ grayscale images of handwritten digits. The EYB database contains frontal colored images of size $192\times 168$ of $38$ individuals. Using the MNIST dataset we pick randomly $35$ images of $4$ randomly selected digits giving rise to a graph with $n=140$ nodes each associated with a  data vector of size $784\times 1$. The approach is repeated $50$ times to perform averaging when testing the performance. In a similar fashion, EYB is used to randomly pick roughly $20$ images for $8$ randomly chosen individuals giving rise to a graph with $n=160$ nodes. Each facial image is rescaled to a size of $32\times 32$ and converted to grayscale, thus here $D=1,024$ entries.

For the MNIST dataset the adjacency matrix $\bbS$ is built such that its $(i,j)$th entry is given as 
%%
%%
%\begin{equation}\label{Eq:MNIST_W}
${\bbx_i^T\bbx_j}\cdot({\|\bbx_i\|\|\bbx_j\|})^{-1}, $
%\end{equation}
%%
%%
whereas for the EYB a Gaussian similarity kernel is employed where $[\bbS]_{i,j}=e^{-0.5\alpha\|\bbx_i-\bbx_j\|_2^2}$ and 
$\alpha=0.01$. A k-nearest neighbor rule is applied where for each node connectivity with the $k=12$ most similar neighbors is preserved.% while for the remaining ones the weights are set to zero in $\bbS$.

\begin{figure}[htp]
	\centering
	\includegraphics[scale=0.26]{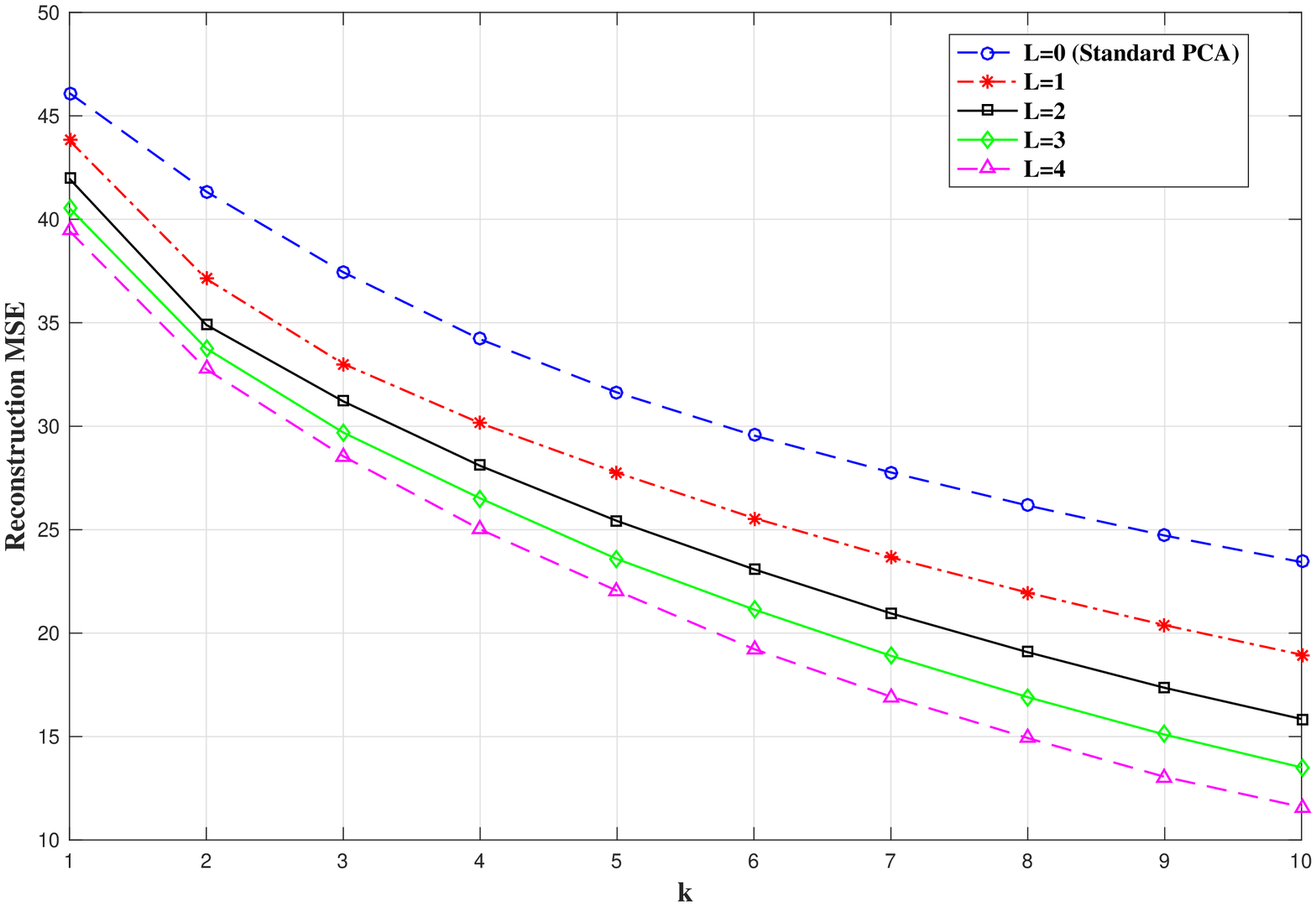}
	\vspace{-10pt}
	\caption{Reconstruction MSE versus $k$ in MNIST.} \label{Fig:1}
	\vspace{-25pt}
\end{figure}

\begin{figure}[htp]
	\centering
	\includegraphics[scale=0.26]{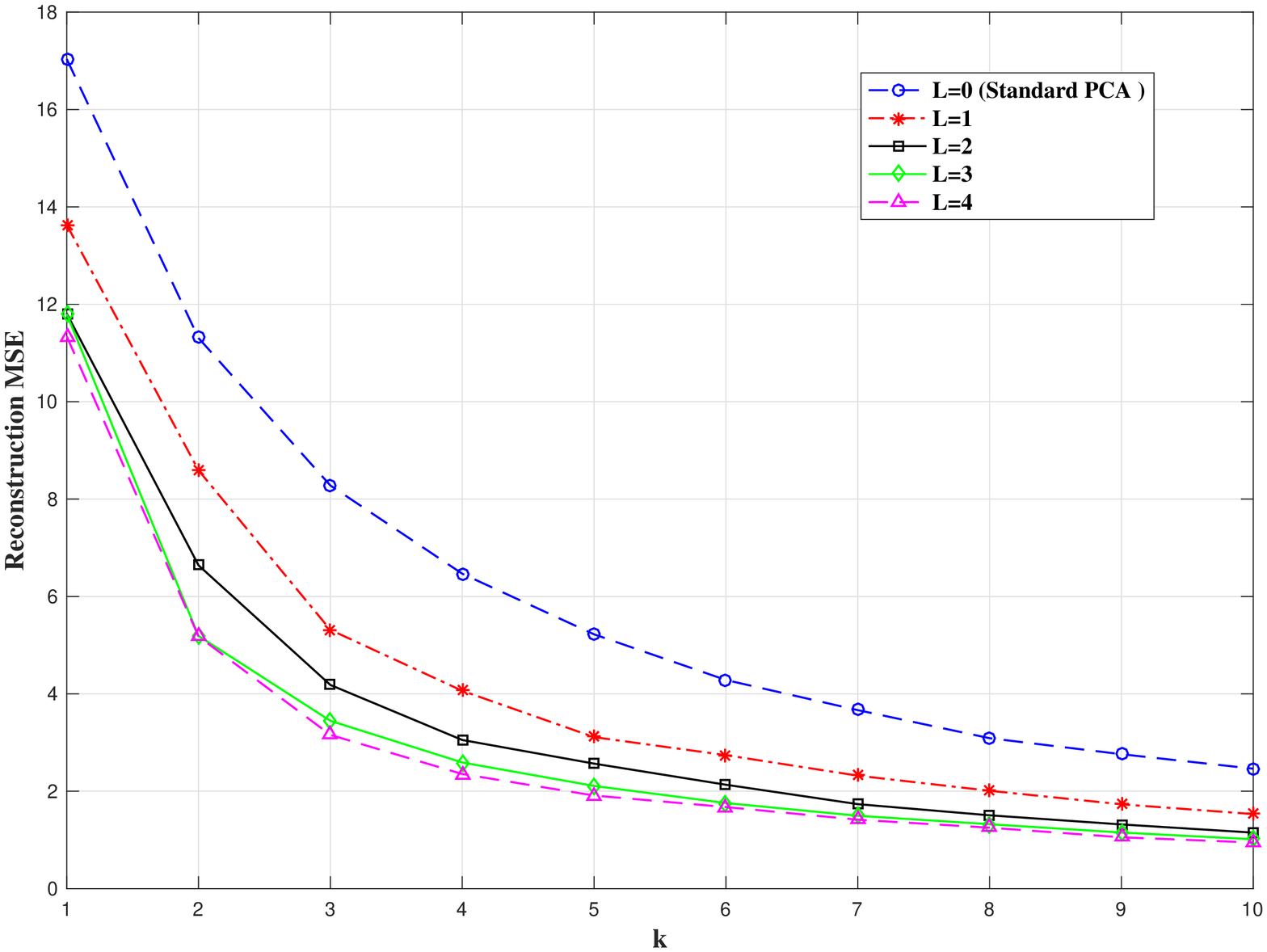}
	\vspace{-12pt}
	\caption{Reconstruction MSE versus $k$ in EYB.} \label{Fig:2}
	\vspace{-15pt}
\end{figure}

Fig. \ref{Fig:1} depicts the reconstruction MSE, in the MNIST-derived dataset, versus the reduced dimension $k$ for the standard PCA ($L=0$), as well as different graph  matrix filters orders $L=1,2,3$ and $4$.  Clearly,
the introduction of graph filtering leads to much lower reconstruction MSE which improves as $L$ increases. Though, after a certain filter order the MSE reduction becomes negligible.  Similar conclusions can be drawn from Fig. \ref{Fig:2} that depicts the  
reconstruction MSE associated with the EYB-derived dataset. The utilization of similarity information in the adjacency matrix of the graph boosts the reconstruction performance over PCA ($L=0$).

\section{Conclusion}
\vspace{-0.3cm}
A novel graph-filtering based data reduction and reconstruction scheme was proposed. A novel formulation incorporates in the reconstruction MSE  graph-filtering, that takes into account data vector similarities. Working in the graph spectral domain enables the derivation of computationally efficient gradient descent techniques to determine the reducing and reconstruction matrix filters. Numerical tests on the image datasets EYB and MNIST demonstrate the improvement in reconstruction quality with respect to standard PCA.

%\vfill\pagebreak

%\section{REFERENCES}
%\label{sec:refs}
\nocite{*}
%\newpage
% References should be produced using the bibtex program from suitable
% BiBTeX files (here: strings, refs, manuals). The IEEEbib.bst bibliography
% style file from IEEE produces unsorted bibliography list.
% -------------------------------------------------------------------------
\bibliographystyle{IEEEbib}

\begin{thebibliography}{1}

\itemsep -5pt
\bibitem{Sampling_1}
A. Anis, A. Gadde, and A. Ortega, ``Towards a Sampling Theorem for Signals on Arbitrary Graphs,'' \emph{Proc. IEEE Int. Conf. Acoust., Speech, Signal Process. (ICASSP)}, pp. 3864--3868, May 2014.

\bibitem{Bertsekas_Nonlinear_Book}
D.~P. Bertsekas,
\newblock {\em Nonlinear {P}rogramming},
\newblock Second Edition, Athena Scientific, 2003.


\bibitem{Brillinger}
D. R. Brillinger, \emph{Time Series: Data Analysis and Theory}. Expanded
Edition, Holden Day, 1981.

\bibitem{Graph_NMF}
D. Cai, X. He, J. Han, and T. S. Huang, ``Graph Regularized Nonnegative Matrix Factorization for Data Representation,'' \emph{IEEE Trans. Pattern Anal. Mach. Intell.,} vol. 33, no. 8, pp. 1548--1560, Aug. 2011.

\bibitem{Sampling_2}
S. Chen, R. Varma, A. Sandryhaila, and J. Kovacevic, ``Discrete Signal Processing on Graphs: Sampling Theory,'' \emph{IEEE Trans. Signal Process.,} vol. 63, no. 24, pp. 6510--6523, Dec. 2015.

%\bibitem{Horn_Johnson}
%R. A. Horn and C. R. Johnson, Matrix Analysis. Cambridge, U.K.:
%Cambridge Univ. Press, 1985.

\bibitem{Graph_Lapl_PCA}
B. Jiang, C. Ding, and J. Tang, ``Graph-Laplacian PCA: Closed-form Solution and Robustness,'' \emph{in Proc. IEEE Conf. Comput. Vis. Pattern Recog. (CVPR)},  2013, pp. 3492--3498.

\bibitem{Multiple_Graph_Coding}
T. Jin, Z. Yu, L. Li, and C. Li, ``Multiple Graph Regularized Sparse
Coding and Multiple Hypergraph Regularized Sparse Coding for Image
Representation,'' \emph{Elsevier Neurocomputing}, vol. 154, pp. 245--256, 2015.

%\bibitem{Estimation_Theory}
%S.~M. Kay,
%\newblock {\em {F}undamental of {S}tatistical {S}ignal {P}rocessing:
%{E}stimation {T}heory},
%\newblock Prentice Hall, 1993.

\bibitem{EYB}
K. C. Lee, J. Ho, and D. J. Kriegman, ``Acquiring Linear Subspaces for Face Recognition Under Variable Lighting,'' \emph{IEEE Trans. Pattern Anal.
Mach. Intell.}, vol. 27, no. 5, pp. 684--698, May 2005.

\bibitem{MNIST}
MNIST Dataset: Available: http://yann.lecun.com/exdb/mnist/


\bibitem{Time_Varying_Graph_Signal_Rec}
K. Qiu, X. Mao, X. Shen, X. Wang, T. Li and Y. Gu, ``Time-Varying Graph Signal Reconstruction,'' \emph{IEEE Journal of Selected Topics in Signal Processing}, vol. 11, no. 6, pp. 870--883, Sept. 2017.


\bibitem{Tutorial_FFT_Filtering}
A. Sandryhaila and J. M. F. Moura, ``Discrete Signal Processing on Graphs,'' \emph{ IEEE Trans. Signal Processing}, vol. 61, no. 7, pp. 1644-1656, 2013.

\bibitem{Graph_Signal_Rec_Percolation}
S. Segarra, A. G. Marques, G. Leus and A. Ribeiro, ``Reconstruction of Graph Signals Through Percolation from Seeding Nodes,''  \emph{IEEE Transactions on Signal Processing}, vol. 64, no. 16, pp. 4363--4378, Aug. 2016.

\bibitem{Robust_PCA_Graphs}
N. Shahid, N. Perraudin, V. Kalofolias, G. Puy, and P. Vandergheynst,
``Fast Robust PCA on Graphs,'' \emph{IEEE J. Sel. Topics Signal Process.,}
vol. 10, no. 4, pp. 740--756, Feb. 2016.

\bibitem{Shen_GG}
Y. Shen, P. A. Traganitis and G. B. Giannakis, ``Nonlinear Dimensionality Reduction on Graphs,'' 
\emph{2017 IEEE 7th International Workshop on Computational Advances in Multi-Sensor Adaptive Processing (CAMSAP)}, 
Curacao, 2017, pp. 1-5.

\bibitem{Graph_Tutorial}
D. I. Shuman, S. K. Narang, P. Frossard, A. Ortega, and P. Vandergheynst,
``The Emerging Field of Signal Processing on Graphs: Extending
High-Dimensional Data Analysis to Networks and Other Irregular
Domains,'' \emph{IEEE Signal Process. Mag.,} vol. 30, no. 3, pp. 83--98, May
2013.

\bibitem{Wang_Sampling}
X. Wang, P. Liu, and Y. Gu,``Local-Set-Based Graph Signal Reconstruction,''  \emph{ IEEE Trans. Signal Processing}, vol. 63, no. 9, pp. 2432--2444, 2015.



\end{thebibliography}

\end{document}